# Marginalizing Out Future Passengers in Group Elevator Control

**Daniel Nikovski and Matthew Brand**
Mitsubishi Electric Research Laboratories
201 Broadway, Cambridge, MA 02139, USA
nikovski@merl.com, brand@merl.com

**Abstract**

Group elevator scheduling is an NP-hard sequential decision-making problem with unbounded state spaces and substantial uncertainty. Decision-theoretic reasoning plays a surprisingly limited role in fielded systems. A new opportunity for probabilistic methods has opened with the recent discovery of a tractable solution for the expected waiting times of all passengers in the building, marginalized over all possible passenger itineraries [Nikovski and Brand, 2003]. Though commercially competitive, this solution does not contemplate future passengers. Yet in up-peak traffic, the effects of future passengers arriving at the lobby and entering elevator cars can dominate all waiting times. We develop a probabilistic model of how these arrivals affect the behavior of elevator cars at the lobby, and demonstrate how this model can be used to very significantly reduce the average waiting time of all passengers.

## 1 INTRODUCTION

Group elevator scheduling is a well-known hard industrial problem characterized by huge state spaces and significant uncertainty [Barney, 2003]. When a new passenger arrives and requests elevator service by pressing a hall-call button, the group controller must assign the passenger to an elevator car with the goal of minimizing his/her waiting time, as well as the waiting times of all existing and future passengers.

The stream of arriving passengers is a stochastic process, which introduces substantial uncertainty in decision making. Each passenger is described by three random variables: time of arrival, floor of arrival, and desired destination floor. All of these variables are sources of uncertainty that must be considered when deciding which car will service a newly arrived passenger.

It is helpful to classify passengers into several groups, according to the type of uncertainty they introduce into the decision-making process:

1. The newly-arrived passenger, whose arrival time and floor are known, but whose destination of travel is not known. In most elevator systems, only the desired *direction* of travel is known, as indicated by pressing one of two hall-call buttons.

2. Existing passengers who have already arrived, but have not boarded a car yet. Like the newly-arrived passenger, their respective arrival times, floors, and desired directions of travel are known, but their exact desired destination floors are uncertain.

3. Future passengers who have not arrived yet. Nothing about such passengers is certain, and only the stochastic parameters of their arrival process are known or can be estimated from data.

An assignment decision influences the waiting times of all three groups of passengers, so the uncertainty introduced by each of them has to be considered.

Ideally, a group elevator controller would compute the marginal costs of all possible assignments with regard to all sources of uncertainty before making a decision. Instead, due to the insurmountable computational complexity of this problem, the vast majority of commercial group elevator schedulers choose to ignore some or all of this uncertainty, typically resorting to heuristic methods.

The earliest schedulers used the simple heuristic principle of *collective group control*, under which a car stops to service the nearest call in its current direction of movement [Strakosch, 1998]. Such scheduling is very sub-optimal, and also very unpredictable. For this reason, collective control is considered unacceptable in several Pacific rim societies including Japan, where social norms dictate that passengers should be notified about which car would pick them up immediately upon requesting service.

Another group of algorithms is based on minimization of the remaining response time (RRT) for each passenger, defined as the time it would take for each existing passenger to be picked up by the car prescribed by the current schedule [Powell and Williams, 1992]. These algorithms focus on minimization of the waiting time of existing passengers only, and ignore altogether the effect of the current assignment on the waiting times of future passengers.

Within the algorithms based on RRT minimization, a further distinction can be made between those that ignore the uncertainty associated with the desired destination floors of existing passengers (ESA, FIM, DLB) [Powell and Williams, 1992, Bao et al., 1994], and those that can properly compute the expected RRT of each passenger with respect to this uncertainty (ESA-DP) [Nikovski and Brand, 2003]. The uncertainty associated with future passengers, however, is an entirely different matter. Properly accounting for the effect of the current decision on the waiting times of all future passengers is an extremely complicated problem, for at least two reasons. First, uncertainty associated with future arrivals is much higher — not only is the exact destination floor unknown for future passengers, but also their arrival times and floors are unknown too. Second, the current decision potentially influences the waiting times of passengers arbitrarily far into the future, which makes the optimization horizon of the problem infinite.

In spite of the computational difficulties, ignoring future passengers often leads to very sub-optimal results. The current assignment affects the future trajectories of the cars, and influences their ability to serve future calls in minimal time. One particularly important situation that exemplifies this effect occurs in up-peak traffic — a regime where most passengers arrive at the lobby of the building and request service to one of the upper floors. Up-peak throughput is typically the limiting factor that determines whether an elevator system is adequate for a building.

Consider the following scenario: A hall call is made somewhere above the lobby, a single car has stopped at the lobby, and the controller decides that this is the optimal car to serve the current call, based only on the projected waiting times of *existing* passengers. If the lobby car is dispatched to serve the new call, the lobby remains uncovered and future passengers arriving there will have to wait much longer than if the car had stayed at the lobby. This short-sighted decision, commonly seen in conventional controller traces, has an especially severe impact in up-peak traffic, since the lobby quickly fills with waiting passengers while the car services the lone caller above.

Suppose, however, that there existed another car above the lobby, which could serve the current hall call almost as fast as the one at the lobby; assigning the new call to it would result in a small short-term loss in the waiting time of the passengers who have just arrived, but this loss is likely to be compensated by a much larger long-term gain in the waiting times of future passengers at the lobby. Thus, a controller that could take into consideration the waiting times of future passengers is likely to have an advantage over a greedy and short-sighted controller that ignores them.

Several methods have been proposed to account for the wait of future passengers, with varying success. Some controllers use fuzzy rules to identify situations similar to the one discussed above and make decisions that are more robust to future events [Ujihara and Tsuji, 1988]. This approach, however, has major disadvantages, such as the need to determine and encode the rules manually, as well as the often unintended manner in which fuzzy-rule inference interpolates between them.

Another approach to accounting for the wait of future passengers has been proposed by Crites and Barto, who recognized that group elevator scheduling is a sequential decision making problem and employed the Q-learning algorithm to asynchronously update the expected costs-to-go (future passengers' waits) of all states of the elevator bank [Crites and Barto, 1998]. They dealt with the huge state space of the system by means of a neural network which compactly approximated the costs-to-go of all states. Their approach is well founded in decision theory and holds significant promise, but its computational demands render it completely impractical for commercial systems. It took 60,000 hours of simulated elevator operation for the algorithm to converge for a single arrival profile, and the resulting reduction of waiting time with respect to other much faster algorithms was only 2.65%, which does not justify its computational costs. Crites and Barto only reported experiments for one down-peak traffic profile and made no comments on other traffic regimes; due to the computational costs, experimentation in varied up-peak regimes is not practical.

In contrast to these labor-intensive and computationally expensive methods, we propose a decision-theoretic approach to choosing the optimal car assignment with respect to both existing and future passengers in up-peak traffic. While it makes some simplifying assumptions of its own, it provides quantitative estimates of the trade-off between waiting times of existing and future passengers, so that a rational scheduling decision can be made. The resulting algorithm is fast and clearly outperforms the state of the art, typically reducing passenger waits by 5% to 55%.

## 2　WAITS OF FUTURE PASSENGERS

In typical up-peak traffic, between 80% and 95% of all future passengers arrive at the lobby. The waiting times of these lobby arrivals is the dominant component in the overall waiting time of future passengers, and the current decision of the scheduler should primarily attempt to minimize the wait at the lobby. Hence, we will begin with the simplifying assumption that *all* future passengers will arrive at the lobby. The effect of unmodelled above-lobby future arrivals will shorten the time-horizon in which predicted waits are accurate; this will be explicitly worked into the calculations later as a discounting factor.

Under the lobby-arrivals-only assumption, it can be seen that the current decision of the scheduler affects the waiting times of future passengers only through the future landing times of cars at the lobby. Calculating these landing times effectively marginalizes out individual future lobby passengers. The optimal strategy to service lobby passengers is to send all cars to the lobby immediately after they have completed servicing their prior commitments to existing passengers. For a building with $C$ shafts, define a *lobby landing pattern* to be an array of times $\mathbf{T} \doteq [T_1, T_2, \cdots, T_C]$, $T_j \geq 0$, where $T_j$ is the arrival time of car $j = 1..C$ at the lobby after it has delivered all of its assigned passengers. Since there is uncertainty about the destinations of passengers currently assigned to a car but not yet boarded, the landing pattern $\mathbf{T}$ is a (vector-valued) random variable with a probability distribution $P(\mathbf{T})$, $\mathbf{T} \in \mathcal{T}$ over the space of all possible landing patterns $\mathcal{T}$.

Ideally, the scheduler should compute the expected waiting time $V(\mathbf{T})$ for each possible landing pattern $\mathbf{T} \in \mathcal{T}$, and take the expectation of that time with respect to the probability distribution $P(\mathbf{T})$:

$$\langle V(\mathbf{T}) \rangle = \int_{\mathbf{T} \in \mathcal{T}} P(\mathbf{T}) V(\mathbf{T}) d\mathbf{T}. \qquad (1)$$

The integral gives the exact estimate of the waiting times of lobby passengers under the lobby-arrivals assumption, but it is not computable because the probability distribution $P(\mathbf{T})$ can only be known through explicit enumeration of all (countably infinite) possible future scenarios via simulation. Even if there were an analytic form for $P(\mathbf{T})$, the size of the (finite) space $\mathcal{T}$ of all possible landing patterns is huge; integrating over it is not practical computationally. Instead, we will use as a substitute the landing pattern consisting of the individual *expected* arrival times of each car $\bar{\mathbf{T}} = [\bar{T}_1, \bar{T}_2, ..., \bar{T}_C] = [\langle T_1 \rangle, \langle T_2 \rangle, ..., \langle T_C \rangle]$, and will employ the approximation $\langle V(\mathbf{T}) \rangle \approx V(\langle \mathbf{T} \rangle) = V(\bar{\mathbf{T}})$. Note that the equality $\langle \mathbf{T} \rangle = \bar{\mathbf{T}}$ is true because each of the components $\mathbf{T}_j$, $j = 1..C$, is an independent random variable whose uncertainty depends only on the probability distribution over the destinations of passengers assigned to car $j$. For the same reason, we may expect the approximation itself to be quite good on average. $V(\bar{\mathbf{T}})$ can be tractably computed using the recently introduced ESA-DP (empty the system via dynamic programming) algorithm [Nikovski and Brand, 2003], which efficiently computes the exact expected arrival time of each car $\bar{T}_i$ (with respect to its current passenger pick-up commitments and their uncertain destinations).

So far we have considered the arrival patterns $\mathbf{T}$ and $\bar{\mathbf{T}}$ as functions of a *fixed* existing assignment of passengers to cars. However, the current decision of the controller—namely to which car the current hall call should be assigned—changes this assignment: Since the controller has a choice between $C$ cars, there are $C$ possible resulting assignments and hence $C$ possible distributions over landing patterns. If we want to employ the approximation discussed above, we need the expected landing pattern $\bar{\mathbf{T}}(i) = [\bar{T}_{i1}, \bar{T}_{i2}, \cdots, \bar{T}_{iC}]$, $i = 1..C$, which would occur if the current call is assigned to car $i$. The meaning of each entry $\bar{T}_{ij}$ is the expected landing time of car $j$ if the current hall call is assigned to car $i$.

Once the matrix of $C$ landing patterns is built, the expected cumulative waiting time of lobby passengers corresponding to each of the $C$ landing patterns (rows of the matrix) can be computed. We now develop a procedure for computing the cumulative waiting time of future lobby passengers as a function of any landing pattern $\mathbf{T} = [T_1, T_2, \cdots, T_C]$.

Since the waiting time of future lobby passengers at the lobby is invariant with respect to the particular order of car arrivals (e.g., it makes no difference whether car 2 arrives in 10 seconds and car 3 arrives in 50 seconds, or vice versa, since both will be empty in up-peak traffic), we assume that the landing pattern $\mathbf{T}$ is already sorted in ascending order: $0 \leq T_1 \leq T_2 \leq \ldots \leq T_C$. Under this assumption, we define $V^0(\mathbf{T})$ to be the expected cumulative waiting time of all future lobby passengers within the time interval $t \in [0, T_C]$:

$$V^0(\mathbf{T}) \doteq \int_0^{T_C} n(t) dt, \qquad (2)$$

where $n(t)$ is the expected number of people waiting to be picked up at the lobby at time $t$.

Before presenting the computational procedure, we will discuss the need to introduce exponential discounting of future waiting times because of a bias in the predicted landing times. The bias is due to our approximating assumption that no future arrivals above the lobby will occur before the end of the landing pattern. In reality, such calls do occur, albeit infrequently; since they have to be accommodated by the cars in service, these cars will be delayed in reaching the lobby. Thus the landing times estimated by the ESA-DP algorithm may underestimate the actual times, very modestly for near-future predictions and significantly for far-future predictions.

A standard method to discount estimates far into the future is to multiply them by $\exp(-\beta t)$, where $\beta > 0$ is a discounting factor [Bertsekas, 2000]. Similarly to the case above, we define the expected *discounted* cumulative waiting time of lobby passengers to be

$$V^\beta(\mathbf{T}) \doteq \int_0^{T_C} e^{-\beta t} n(t) dt. \quad (3)$$

Consider splitting the interval $[0, T_C]$ into $C$ different intervals $[T_{i-1}, T_i]$, $i = 1..C$ (setting $T_0 = 0$). On first consideration, it would seem that the expected number of people waiting at time $t \in [T_{i-1}, T_i]$ is proportional to the time elapsed since the last time a car visited the lobby ($T_{i-1}$). If we model that arrival of lobby passengers as a Poisson process with rate $\lambda$, the expected number of people waiting is simply $n(t) = \lambda(t - T_{i-1})$, and the integral above splits into $C$ easily evaluable parts. (We assume here that cars pick up instantly all people they find waiting at the lobby, since load times are very small relative to wait times).

Unfortunately, this reasoning ignores the fact that if car $i$ reaches the lobby and finds it empty, it will not depart immediately (at its arrival time $T_i$), but will wait at the lobby until the next future passenger arrives and boards. Furthermore, this approach cannot handle a very important special case: If there are already $j$ cars at the lobby at time $t = 0$, the first $j$ passengers will not wait at all — each will immediately board a waiting car and ride up, with little or no waiting time. The significant but speculative savings in this scenario must be balanced against the real cost of not using those cars to service known passengers above the lobby. In order to quantify these savings, we must accurately model the behavior of elevator cars at the lobby.

## 3 A SEMI-MARKOV SYSTEM MODEL

In order to correctly estimate the waiting times of lobby passengers given the actual behavior of cars when they find nobody waiting at the lobby, we employ a semi-Markov chain whose states and transitions describe the behavior of landing lobby cars in response to passenger traffic at the lobby.

A semi-Markov chain consists formally of a finite number of states $S_i$, $i = 1..N_S$, average momentary costs $r_{ij}$, expected transition times $\tau_{ij}$, and probabilities $P_{ij}$ of the transitions between each pair of states $S_i$ and $S_j$, and an initial distribution $\pi(S_i)$ which specifies the probability that the system would start in state $S_i$ [Bertsekas, 2000]. Furthermore, each semi-Markov chain contains an embedded fully-Markov chain evolving in discrete time, whose cumulative transition costs $R_{ij}$ are defined as $R_{ij} \doteq \tau_{ij} r_{ij}$, and all transitions are assumed to occur within a unit of time.

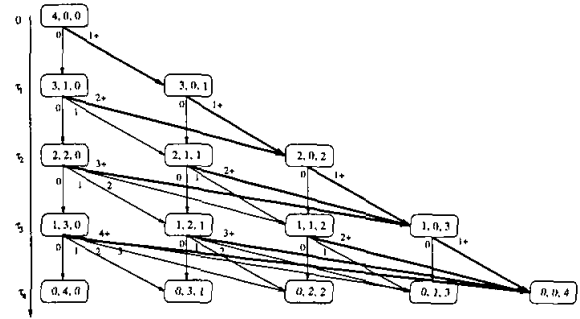

Figure 1: Grid structure for the embedded semi-Markov chain for a building with four shafts. Row $i$ of the model contains all possible states of the system just after car $i$ has arrived at time $T_i$ and has picked up all passengers that might have been waiting at the lobby. (Note that the vertical time axis is not drawn to scale.) Only transitions shown in bold arrows have non-zero costs; the costs of all other transitions are zero. Transitions labeled with $n+$ are taken when $n$ *or more* new passengers arrive.

The states in the semi-Markov chain used for our problem are labeled by the triple $(i, j, m)$, where $i$ is the number of cars that have yet to arrive at the lobby, $j$ is the number of cars currently at the lobby waiting for passengers, and $m = C - i - j$ is the number of cars already departed from the lobby. Accordingly, we place the states of the semi-Markov chain in a two-dimensional grid (matrix), whose element $S_{im}$ corresponds to state $(i, j, m)$ (figure 1). Row $i$ of the model matrix contains all possible states of the system immediately after car $i$ has arrived at time $T_i$ and has picked up all passengers that might have been waiting at the lobby at that time.

We will first provide a solution for the generic situation represented by this model, namely when no cars are present at the lobby at the current decision time $(T_1 > 0)$, and later extend the solution to the case when some cars are currently parked at the lobby. For the generic case, the starting state of the chain is the state $(C, 0, 0)$, i.e., all $C$ cars have yet to arrive at the lobby. The terminal states are those in the bottom row of the model, when all $C$ cars have already landed, and depending on how many passengers have arrived in the interval $t \in [0, T_C]$, either all cars have departed with passengers on board (state $(0, 0, C)$), or some cars are still present at the lobby (states $(0, j, C - j)$ for some $j > 0$).

Each state $(i, j, m)$ in the rows above the bottom one $(i > 0)$, where $j = C - i - m$, can transition to two or more successor states, depending on exactly how many new lobby passengers would arrive during the time interval $t \in [T_i, T_{i+1}]$. For example, the chain would transition from state $(4, 0, 0)$ to state $(3, 1, 0)$ only if no new passengers arrive by time $T_1$, and will transition to state $(3, 0, 1)$ if one or more passengers arrive by that time. Each of the transitions in figure 1 is labeled with the number of passengers that should arrive if this transition is to be taken.

The time to complete each transition is readily determined to be the interval $\Delta T_i = T_i - T_{i-1}$ between two car arrivals. The probability of each transition is also easy to compute, since it is equal to the probability that a particular number of people would arrive within a fixed interval from a Poisson process with arrival rate $\lambda$. Thus, the probability $p(x)$ that exactly $x$ people would arrive in time $\Delta T_i$ is $p(x) = (\lambda \Delta T_i)^x e^{-\lambda \Delta T_i}/x!$. For transitions labeled with an exact number of arriving passengers, this formula can be used directly. For transitions labeled with $n+$, meaning that they are taken when $n$ or more new passengers arrive, the probability of the transition is the complement to one of the sum of the probabilities of all remaining outgoing transitions from this state: $p(n+) = 1 - \sum_{x=0}^{n-1} p(x)$.

Computing the cost of transitions labeled with an exact number of passengers is trivial: Since the number of arriving passengers is less than or equal to the number of cars available at the lobby, none of these passengers would have to wait and the cost of the corresponding transitions is zero. Computing the cost of the last (rightmost) transition from each state, shown in bold in figure 1, however, is quite involved. Such a transition corresponds to the case when $n$ or more people would arrive at the lobby, while only $n-1$ cars are present there. The computation has to account for the fact that if $x$ new passengers arrive, and $x \geq n$, the first $n-1$ of them would each take a car and depart without waiting, and only the remaining $x - n + 1$ people would have to wait.

Figure 1 shows that for any state $S_{im}$ of the grid, as defined above, and $j = C - i - m$, the transition shown in bold is taken when more than $j$ people arrive, i.e., $n = j - 1$. Hence, if that transition is taken and $x$ new passengers arrive, only the last $x - j$ of them would have to wait. In other words, if $x$ passengers have appeared within some time $t$, the differential (momentary) cost $r_{im}$ at that time would be $x - j$.

Since such a transition covers the cases when some number of passengers greater than $j$ would appear, and this number could theoretically be arbitrarily large even in a finite time interval, the expected cost of the transition would be a weighted sum over all possible numbers of arrivals $x$, from $j + 1$ to infinity, and the weights would be the probabilities that $x$ arrivals would occur, as given by the Poisson distribution. In addition, the differential costs at time $t$ should be discounted by a factor of $\exp(-\beta t)$, as discussed previously. This reasoning yields the following expression for the expected discounted cumulative waiting time $R_{im}^\beta$ of lobby passengers during the last transition out of state $S_{im}$, with $j = C - i - m$:

$$R_{im}^\beta = \int_{T_{C-i}}^{T_{C-i+1}} e^{-\beta t} \sum_{x=j+1}^{\infty} \frac{[\lambda(t - T_{C-i})]^x e^{-\lambda(t-T_{C-i})}}{x!} (x-j) dt. \quad (4)$$

After a change of integration variables, simplification, and splitting of the integral into two parts according to the two components of the difference $x - j$, the expression for the cost evaluates to $R_{im}^\beta = e^{-\beta T_{C-i}}[F(\Delta T_{C-i+1}) - F(0)]$, where we make use of the function

$$\begin{aligned} F(t) &= \sum_{x=0}^{j} \lambda^x e^{-(\lambda+\beta)t}(x-j) \sum_{l=0}^{x} \frac{t^{x-l}}{(x-l)!(\lambda+\beta)^{l+1}} \\ &+ \frac{(\beta j - \beta \lambda t - \lambda)e^{-\beta t}}{\beta^2} + c_0 \end{aligned} \quad (5)$$

for some arbitrary, but fixed integration constant $c_0$, which we set to zero for computational convenience. Certainly, the above function is valid only when at least some discounting is used ($\beta > 0$); when $\beta = 0$, the cost evaluates to

$$R_{im}^0 = G(\Delta T_{C-i+1}) - G(0), \quad (6)$$

for

$$G(t) = \sum_{x=0}^{j} \lambda^x e^{-\lambda t}(x-j) \sum_{l=0}^{x} \frac{t^{x-l}}{(x-l)!\lambda^{l+1}} + \frac{\lambda}{2}t^2 - jt + c_0. \quad (7)$$

Once all costs and probabilities of the semi-Markov model have been computed as described above, the cumulative cost (wait) incurred by the system if it starts in any of the model states can be computed efficiently by means of dynamic programming, starting from the bottom row of the model and working upwards. Since the states in the bottom row are terminal and mark the end of the landing pattern, we set their costs-to-go to zero, i.e., we are not interested in the amount of passenger wait accumulated after the last landing.

Once the costs-to-go of all states are known, we can read off the cumulative waiting time for the whole landing pattern $\mathbf{T}$ from the initial state of the model. In the generic case, when no cars are present at the lobby at time $t = 0$, the initial state is always $(C, 0, 0)$. The special case when $l$ cars are present at the lobby at time $t = 0$ can be handled just as easily — in this case, the starting state is $(C - l, l, 0)$, and the expected discounted cumulative wait for the whole landing pattern is the cost-to-go of this starting state $(S_{C-l,0})$. This eliminates the need to handle this special case separately from the generic one.

## 4 COMBINING ESTIMATES

The algorithm described above provides estimates $V_i^\beta \doteq V^\beta(\mathbf{T}_i)$ of the expected cumulative discounted waiting time of *future* lobby passengers, based on each of the $C$ landing patterns $\mathbf{T}_i$ resulting from the decision to assign the current hall call to car $i$, $i = 1..C$. Simultaneously, the ESA-DP algorithm [Nikovski and Brand, 2003] gives exact estimates $W_i$ of the cumulative non-discounted waiting time of all *existing* passengers of all cars, including the one(s) that signaled the current hall call, if this call is assigned to car $i$, $i = 1..C$. In order to arrive at an optimal decision balancing the wait of both existing and future passengers, the two sets of values $V_i^\beta$ and $W_i$ have to be combined in an appropriate manner.

There are significant differences between these two measures: The cumulative waiting time of existing passengers $W_i$ is not discounted, while the cumulative waiting time of future passengers is discounted. Furthermore, the objective of the scheduling algorithm is to minimize the *average* waiting time, and not the *cumulative* waiting time over some interval — the two measures are interchangeable for the purposes of optimization only when the time intervals for all possible decisions are equal. Since this is not the case (in general, the landing patterns for different cars do not have the same duration), the scheduling algorithm would have to obtain average waiting times from their cumulative counterparts.

Obtaining the average waiting time of existing passengers $\overline{W}_i$ from the cumulative waiting time $W_i$ is trivial — the number $N$ of currently waiting passengers is always known by the controller and does not depend on the candidate car number $i$, so $\overline{W}_i = W_i/N$. On the other hand, obtaining the average waiting time of future passengers $\overline{V}_i$ from the cumulative discounted waiting time $V_i^\beta$ over the duration of a landing pattern is not as obvious. The duration $T_C$ of the landing pattern is known, and if the arrival rate at the lobby is $\lambda$, the expected number of arrivals within $T_C$ time units is $\lambda T_C$. However, dividing $V_i$ by $\lambda T_C$ is meaningless, because $V_i$ has been discounted at a discount rate $\beta$.

Instead, we can think of the discount factor $\exp(-\beta t)$ as an averaging weight for time $t$. If $n(t)$ is the expected momentary number of people waiting at time $t$ as reflected in the costs of the Markov model,

$$V_i^\beta = \int_0^{T_C} e^{-\beta t} n(t) dt \qquad (8)$$

has the meaning of expected *cumulative* weighted number of people waiting during the interval $[0, T_C]$. Therefore the quantity

$$\bar{n} = \int_0^{T_C} e^{-\beta t} n(t) dt / \int_0^{T_C} e^{-\beta t} dt \qquad (9)$$

is the expected *average* number of waiting people within this interval, properly normalized by the sum (integral) of all weight factors. Furthermore, Little's law specifies that $\bar{n} = \lambda \overline{V}_i$ [Cassandras and Lafortune, 1999], which finally yields the time-normalized expected wait of future passengers:

$$\overline{V}_i = V_i^\beta \beta / (\lambda - \lambda e^{-\beta t}). \qquad (10)$$

Having obtained comparable estimates $\overline{W}_i$ and $\overline{V}_i$ of the waiting times of existing and future passengers, they have to be combined into a single performance criterion, for example by means of a single weight $0 \leq \alpha \leq 1$, such that the performance criterion would be $\alpha \overline{W}_i + (1-\alpha)\overline{V}_i$. The balance between present and future waits depends on how quickly the system can free itself of present constraints by delivering passengers. Thus the optimal value of $\alpha$ is ultimately an empirical question, depending mostly on the physical performance of the elevator system. In our experiments, we found that values of $\alpha$ within the interval $[0.1, 0.3]$ stably produced the best results, regardless of the height of the building and number of shafts.

The resulting algorithm, which uses a weighted average of ESA-DP's estimates $\overline{W}_i$ and the look-ahead estimates $\overline{V}_i$, will be called ESA-DP-LA (ESA-DP with Look-Ahead). It is invoked at each passenger's arrival, and its only parameter is the current arrival rate $\lambda$, of which empirical estimates are computed and maintained in most modern elevator scheduling systems [Amano and Masude, 2002]. The complexity of evaluating the look-ahead estimates $\overline{V}_i$ is $O(C^2)$, and since the number of cars $C$ is always small, the computational time for producing these estimates is negligible with respect to that necessary for computing the expected waits of current passengers $\overline{W}_i$ and expected landed patterns $\mathbf{T}_i$ by means of the original ESA-DP algorithm.

## 5 EXPERIMENTS

The ESA-DP-LA algorithm was compared to a conventional method for supervisory group control in a detailed simulator. The conventional controller's basic strategy is to identify a likely path for each car given its commitments, then make a new passenger-to-car assignment that minimizes the round-trip time of all cars along their likely itineraries. Recently fielded systems by a number of market-leading manufacturers generally operate on the same principle [Barney, 2003], although the matter is partly shrouded by trade secrets.

The algorithms were tested on various buildings with height of 8, 15, 20, and 30 floors, served by either 3, 4, 5, 6, 7, or 8 elevator shafts, whose cars were moving at a speed of 3 m/s. Each floor in these buildings was 4m tall, except for the lobby, which was 5m tall.

Each trial consists of a 1 hour simulation with passenger traffic of randomly generated traffic, using a unique random seed. Both algorithms see the same exact traffic. The performance of the two algorithms was tested under arrival rates ranging from 100 arrivals per hour up to the point where average waiting time exceeded one minute. Such a point is reached at different rates for different buildings and number of shafts in the elevator group.

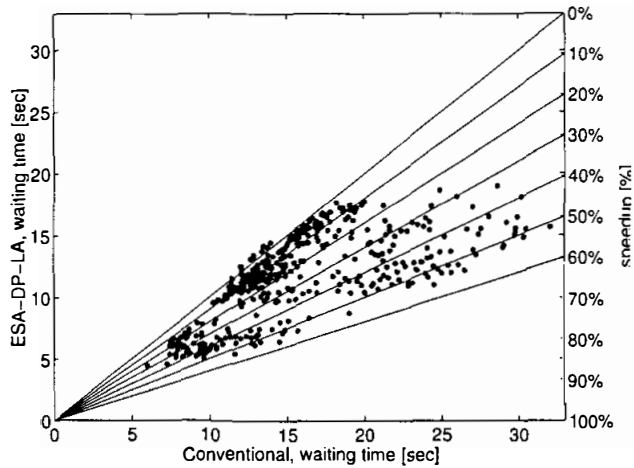

Figure 2: Waiting times of the ESA-DP-LA scheduler plotted against waiting times of the conventional scheduler in identical scenarios, in seconds. Each dot represents an average over 50 hours of simulation in a specific building type and arrival rate. Dots below the diagonal represent cases when ESA-DP-LA achieves lower waiting time than the conventional scheduler, and vice versa for dots above the diagonal. The right axis (speed-up) shows percentage reduction in waiting times.

The experiments explored the case of mixed up-peak traffic. In office buildings this is the most demanding traffic regime, combining maximal arrival rates and uncertainty in passenger destinations. Most (80%) of the traffic originated at the lobby and was directed approximately evenly to the upper floors, while the remaining 20% of the traffic was between floors other than the lobby. The results are plotted in figure 2, and indicate that the algorithm significantly reduces waiting time with respect to the conventional algorithm, with savings in the range of 5%-55%.

As noted, the ESA-DP-LA algorithm is parametrized by the mixing coefficient $\alpha$ and the discounting rate $\beta$. Their values were determined experimentally for each building type in "fitting trials" generated from one set of random seeds. We took care to ensure that the "test trials" graphed in figure 2 are generated from a different set of seeds. In general, the performance of ESA-DP-LA on the two sets of seeds is similar—average wait times differ by roughly one second. Performance is also robust to ±50% changes in $\alpha$ and $\beta$, but larger changes can add more than 10 seconds to the average ESA-DP-LA wait time.

This robustness is illustrated in figure 3 for one specific building (15 floors and 6 shafts). The graph in figure 3 depicts the experimental dependency of passenger waiting times on the value of the discounting parameter $\beta$, for a fixed value of the mixing paramter $\alpha = 0.2$, and two separate sets of 49 random-number seeds. Overall, for both sets, the minimum waiting times tend to be achieved in the same specific interval $0.015 < \beta < 0.025$. Values much less than $\beta = 0.015$ clearly result in poor performance — if too little or no discounting is used, waiting times can increase by up to 7 seconds for this particular building, and by more for other buildings. This can be attributed to the relatively fast rate at which landing-pattern estimates become imprecise. When larger discounting rates are used ($\beta > 0.025$), performance worsens as well, but at a slower rate. (Ultimately, for very large values of the discounting parameter, e.g. $\beta > 2$, the decisions of ESA-DP-LA become identical to those of ESA-DP.) This shows that it is safer to err in the direction of more discounting than in the direction of less discounting.

Figure 3 also shows that, in general, it is not possible to find experimentally the exact "best" value for the discounting parameter. The optimization surface is fairly noisy even when relatively many random-number seeds are used, and gradient-descent search is very hard to apply. Furthermore, waiting times between fitting and testing trials differ by approximately one second, so it is not feasible to achieve better accuracy than that. What is important in practical terms is that this residual variation is much smaller than the overall improvements in waiting times achieved by ESA-DP-LA with respect to ESA-DP. This effect is shown is Figure 4, where waiting times of ESA-DP-LA are plotted vs. those of ESA-DP. Improvements are smaller than when compared to a conventional scheduler, reflecting the advantage ESA-DP already had with respect to the conventional algorithms, but they are still very significant. More importantly, these improvements can be attributed only to the look-ahead policy.

It is also instructive to interpret the experimentally obtained value for the discounting parameter $\beta$ as a measure of how far into the future the scheduler looks ahead. Since $\tau = 1/\beta$ is the time when future waits are discounted $e$ times, the value of $\tau$ can be assumed to be the effective horizon of the scheduler. The interval $0.015 < \beta < 0.025$ corresponds to horizon of $40 < \tau < 67$ seconds, which is on the order of one round trip of an elevator car. This is consistent with our expectation that once a car initially at the lobby has been able to complete its round trip and return to the lobby, the actual landing pattern from then on is very different from our estimates and they should not be relied upon. Furthermore, we can compare our experimental optimal discounting rate with that used by [Crites and Barto, 1998]. They used $\beta = 0.01$, or equivalently $\tau = 100$ seconds — although on the same order as our results, an effective horizon of 100 seconds is probably too long for all but the tallest buildings.

Elevator performance in up-peak traffic typically determines the number of shafts a building will need. Using standard guidelines for elevatoring a building according to the waiting times the system should deliver [Barney, 2003], we estimate that if the industry shifted from current controller technologies to ESA-DP-LA, 10-15% of all new mid- and high-rise office buildings could be built with one less shaft than currently recommended *and* provide superior service.

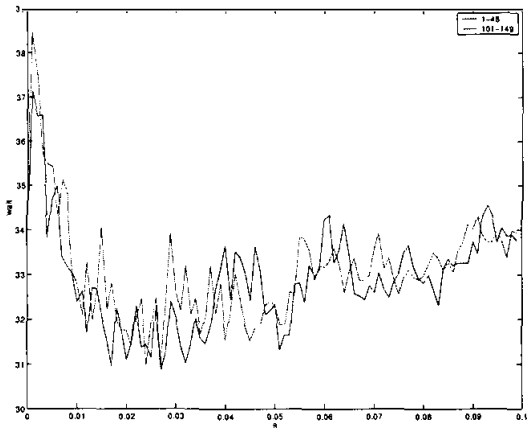

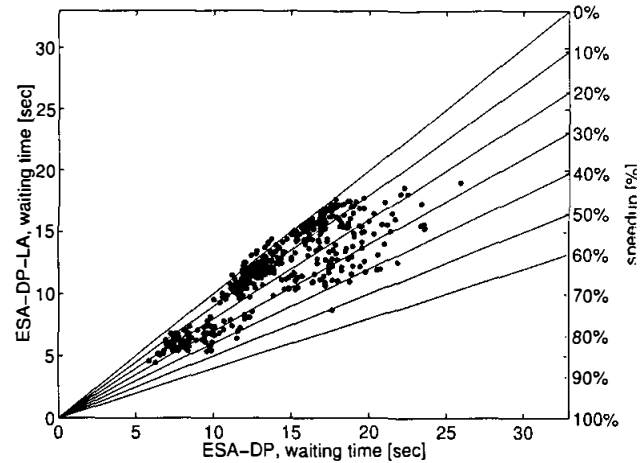

Figure 3: Waiting times of the ESA-DP-LA scheduler plotted against the discount parameter $\beta$, for a fixed mixing coefficient $\alpha = 0.2$, in a 15 floor, 6 shaft building, $\lambda = 2500$. Average values for two sets of 49 random seeds are shown. Both sets show a general minimum in the interval $0.015 < \beta < 0.025$, although the dependency is very noisy locally.

Figure 4: Waiting times of the ESA-DP-LA scheduler plotted against waiting times of ESA-DP in identical scenarios, in seconds.

## 6 SUMMARY

This paper presented an algorithm for approximate estimation of the waiting times of future lobby passengers for each possible assignment available to a scheduling algorithm. We combine an estimator of elevator landing times and a semi-Markov model of overall system behavior to compute the expected waits of future passengers arriving at the lobby. This estimate complements the estimates for the waiting time of passengers already known to the system, and allows the scheduler to make a rational assignment based on the balance between waiting times of existing and future passengers. The resulting scheduling algorithm achieves large improvements in average waiting time of passengers — sometimes halving it or better — and creates real possibilities for reducing the number of shafts required for properly elevatoring a building.